\documentclass[conference]{IEEEtran}
\IEEEoverridecommandlockouts
\usepackage{cite}
\usepackage{amsmath,amssymb,amsfonts}
\usepackage{algorithmic}
\usepackage{graphicx}
\usepackage{textcomp}
\usepackage{xcolor}
\usepackage{url}
\usepackage{comment}
\usepackage{booktabs}
\usepackage{float}

\def\BibTeX{{\rm B\kern-.05em{\sc i\kern-.025em b}\kern-.08em
    T\kern-.1667em\lower.7ex\hbox{E}\kern-.125emX}}
\begin{document}

\title{Behavioral Cloning Models Reality Check for Autonomous Driving\\
\thanks{Identify applicable funding agency here. If none, delete this.}
}

\author{\IEEEauthorblockN{1\textsuperscript{st} Mustafa Yildirim}
\and
\IEEEauthorblockN{2\textsuperscript{nd} Barkin Dagda}
\and
\IEEEauthorblockN{3\textsuperscript{rd} Vinal Asodia}
\and
\IEEEauthorblockN{4\textsuperscript{th} Saber Fallah}

}

\maketitle

\begin{abstract}
How effective are recent advancements in autonomous vehicle perception systems when applied to real-world autonomous vehicle control? While numerous vision-based autonomous vehicle systems have been trained and evaluated in simulated environments, there is a notable lack of real-world validation for these systems. This paper addresses this gap by presenting the real-world validation of state-of-the-art perception systems that utilize Behavior Cloning (BC) for lateral control, processing raw image data to predict steering commands. The dataset was collected using a scaled research vehicle and tested on various track setups. Experimental results demonstrate that these methods predict steering angles with low error margins in real-time, indicating promising potential for real-world applications.

\end{abstract}

\begin{IEEEkeywords}
Behavior Cloning, Autonomous Driving, QCar, Transformers
\end{IEEEkeywords}

\section{Introduction}

End-to-end Autonomous Driving (AD) involves using camera images to control a vehicle's lateral and longitudinal movements. A critical component of end-to-end AD is vision-based steering angle estimation, where the system predicts the vehicle's steering angle using images from an onboard camera system. This task faces significant challenges due to variations in road scene appearance, environmental conditions, illumination, and the non-linear dynamics of vehicles \cite{john2017automated}. Despite extensive research and advancements in neural network architectures and methodologies, there remains an ongoing debate about the most effective approaches to enhance this technology's robustness and reliability \cite{john2017automated}. For successful real-world deployment of autonomous vehicles, it is crucial to develop models that can robustly handle diverse driving conditions and provide accurate and reliable steering predictions.

Behavior Cloning (BC) has emerged as a popular method for end-to-end AD due to its simplicity and effectiveness in tasks like lane following. However, BC models heavily depend on their training datasets, which makes them vulnerable to compounding errors when encountering unfamiliar states. For instance, when a BC model is exposed to a scenario not seen during training, it often mimics actions without understanding the underlying intent, leading to potentially unsafe driving behaviors \cite{hussein2017imitation}. Early work by Pomerleau (1988) \cite{pomerleau1988alvinn} on BC for lane following used a camera-equipped vehicle to collect data for training a neural network-based navigation model. However, this method's effectiveness is limited by the quality of the training dataset, and it struggles to recover from deviations caused by sensor inaccuracies or noise. Nvidia researchers later developed more sophisticated deep learning CNN models \cite{bojarski2016end} to address these issues, but their real-world performance remains underexplored, highlighting the "simulation-to-reality" gap that persists in this field.

Despite advancements in simulation-based testing of BC models using end-to-end CNN algorithms for steering control \cite{farag2019cloning}, and other approaches employing BC as an image classification task \cite{haji2019self}, these studies do not fully capture real-world driving complexities. Therefore, there is a critical need for more comprehensive testing and validation of these methods under realistic conditions to better understand their performance and limitations.

This paper addresses these gaps by evaluating the performance of state-of-the-art perception systems using different architectures, such as Autoencoder-based Behavioral Cloning (AutoBC), Vision Transformers (ViT), and Spatial Attention mechanisms, in real-world settings. We propose a novel framework that integrates these methods and assesses their performance across various track setups collected using a scaled research vehicle. Our study provides a detailed analysis of each method's strengths and weaknesses, contributing to the broader understanding of their applicability in real-world autonomous driving scenarios.

\subsection{Related Work}

Recent advancements in autonomous driving have explored various methods to improve the robustness and generalization of Behavioral Cloning (BC) models. While BC based on Convolutional Neural Networks (CNNs) is a popular choice, it suffers from drawbacks such as compounding errors and poor performance on unseen data. To address these issues, several methods have been proposed, including the DAgger framework \cite{ross2011reduction}, which combines offline training with online learning to reduce compounding errors and improve generalization.

Human-inspired attention mechanisms\cite{cultrera2023explaining} that dynamically adjust focus based on situational familiarity, similar to how drivers prioritize potentially dangerous situations, have been explored in autonomous driving research \cite{gou2022driver}. Additionally, methods using feature extractors and gating mechanisms have shown promise in optimizing sensor input to reduce computational inference time \cite{fang2020multi}. Furthermore, another study \cite{cultrera2020explaining} emphasizes the benefits of visual attention in enhancing model interpretability and decision-making. ViT \cite{cultrera2023addressing} is also implemented with multi-task learning to address key limitations in state-aware imitation learning, such as compounding errors and offline-online performance gaps. By leveraging state token propagation, these models improve performance handling of unseen states and low inertia problems. All these studies employed very large dataset for training comparing to our data samples and performed in simulation platforms. While in real world settings data collection and real time testing has not been tested in these studies.

Rule-based systems, such as those implemented on scaled autonomous vehicles like the QCar \cite{Qcar_product}, have been employed for specific tasks like line following and lane-keeping. Various approaches have been proposed to tackle lane-keeping tasks. For instance, a study \cite{diab2020self} proposed a lane-keeping assist (LKA) system based on PID and Pure Pursuit controllers, integrated with classical lane detection techniques.While these methods prove effective in controlled and predictable environments, they are limited in their ability to handle complex, unstructured, real-world driving scenarios. The main drawback of these systems is their reliance on simple controllers that primarily process RGB colours and measure distances, making them unsuitable for dynamic and unpredictable settings. To address these limitations, more recent approaches have explored knowledge-based models, particularly the integration of Large Language Models (LLMs) \cite{yildirim2024highwayllm}, to enhance situational awareness and decision-making in autonomous vehicles. Although LLMs show potential for processing rule-based information and making context-aware decisions, their effectiveness is contingent on the quality and comprehensiveness of the underlying knowledge base, which remains a significant challenge.

Research has increasingly focused on leveraging deep learning techniques like 3D Convolutional Neural Networks (CNNs) and Recurrent Neural Networks (RNNs) for steering angle prediction in self-driving vehicles. For example, this study \cite{du2019self} proposed a model combining 3D convolutional layers and Long Short-Term Memory (LSTM) networks to capture both spatial and temporal information from sequential images. Although their model achieved promising results, it was computationally intensive, limiting its practicality in real-time systems. This reliance on heavy computational resources and complex architectures presents a challenge for deployment on less powerful hardware, such as scaled research vehicles or embedded systems. Similarly, while transfer learning methods like ResNet with frozen layers provide strong baselines, they do not adequately address real-time adaptation and generalization to unseen environments.

Another study \cite{codevilla2018end} proposes imitation learning methods that train models on human driving demonstrations using high-level commands (left, right, and straight) to direct vehicles toward specific goals or through intersections. However, these approaches face challenges, including the need for larger datasets and the integration of natural language processing for vehicle guidance. Transfer learning has also been proposed as an efficient method for autonomous driving, where neural networks pre-trained on different tasks are fine-tuned with additional layers specific to a new use case. Autoencoder models, for example, are used to initialize models that are then fine-tuned for tasks such as object detection, path prediction, or motion planning \cite{chen2016variational}. However, recent studies have noted limitations; for instance, the authors of \cite{pak2022carnet} did not compare these models with other approaches and treated the problem within a fixed set of steering angles and acceleration values, a significant limitation given the inherently continuous nature of steering control.

The limitation of BC is deeply analysed in a study \cite{codevilla2019exploring} that highlights significant issues such as dataset bias, high variance in policy performance, and difficulties in handling dynamic objects in driving environments for imitation learning. The study thoroughly evaluates BC models in dynamic traffic scenarios and complex urban settings. Their improved BC model incorporates deeper architectures to mitigate some of these challenges in the Carla simulation platform. However, these improvements still do not fully address the need for more adaptive and generalisable models, even in simulation. Our study explores these issues in a real-world setting, with fewer data samples on an embedded vehicle.

While simulation-based approaches offer valuable insights for autonomous driving model development, there remains a significant 'simulation-to-reality' gap. Recent studies highlight that models validated in simulated environments often fail to perform as expected in real-world conditions due to discrepancies in visual, environmental, and behavioral factors \cite{stocco2022mind}. Moreover, a study \cite{codevilla2018offline} demonstrated a significant discrepancy between offline error metrics and online performance, underscoring the limitations of relying solely on offline evaluations. While they propose alternative offline metrics, the real-world applicability of these models remains uncertain. This gap underscores the need for extensive real-world testing to uncover potential failures and ensure robust model performance under realistic conditions. Our study directly addresses this need by evaluating state-of-the-art models in real-world settings, thereby bridging the gap between offline evaluation and actual driving performance and providing a more comprehensive understanding of their applicability and limitations.

While various alternative approaches to BC, including rule-based systems, knowledge-based models, attention mechanisms, and transfer learning, show potential, each has limitations that must be addressed to improve their applicability to real-world autonomous driving scenarios. This paper revisits the current challenges and critically explores how integrating newer techniques, such as autoencoders and transformer encoder-only models, can address existing drawbacks and potentially advance the field.

\section{Methods}

\subsection{Behaviour Cloning}\label{BC}

Behavior Cloning is a supervised learning approach where a policy is learned from a dataset of expert demonstrations. It is used to map raw sensory inputs (like camera images) directly to control commands (e.g., steering angles). The dataset \( D = \{(s_{i}, a_{i})\} \) consists of state-action pairs, where \( s_{i} \) represents the state (e.g., an image), and \( a_{i} \) denotes the corresponding action (e.g., a steering command). The objective of BC is to learn a policy \( \pi^* \) that minimizes the discrepancy between the expert's actions and the learner's actions, defined as:

\[
\pi^* = \arg \min D(q(\phi), p(\phi))
\]

where \( q(\phi) \) is the distribution of the expert policy, \( p(\phi) \) is the distribution of the learner policy, and \( D(q, p) \) measures the divergence between these distributions.

\begin{figure}[htbp]
\centerline{\includegraphics[width=0.49\textwidth,height=\textheight,keepaspectratio]{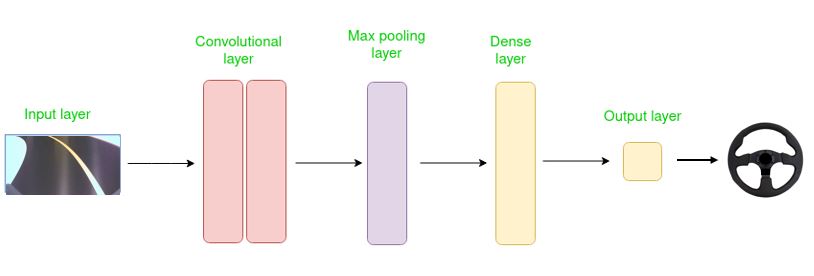}}
\caption{CNN architecture}
\label{CNN_nvidia}
\end{figure}

\subsection{Autoencoder}

We have implemented an autoencoder-based Behavioral Cloning (BC) network to learn compact representations of input images and assess the network's ability to reconstruct these images \cite{chen2016variational}. An autoencoder consists of two main components: an encoder and a decoder. The encoder network compresses the input data into a lower-dimensional latent space, while the decoder reconstructs the input data from this compressed representation.

The encoder network \( f_{\theta} \) consists of three convolutional layers with Exponential Linear Unit (ELU) activation functions and batch normalization. Each convolutional layer is followed by MaxPooling \cite{boureau2010theoretical} for downsampling. The encoder transforms an input image \( x \) into a latent representation \( z \):

\[
z = f_{\theta}(x)
\]

where \( z \) is the latent representation in a lower-dimensional space.

The decoder network \( g_{\phi} \) contains three convolutional layers with ELU activation functions and batch normalization. Each convolutional layer is followed by UpSampling to restore the original image dimensions. The decoder takes the latent representation \( z \) and reconstructs the image \( \hat{x} \):

\[
\hat{x} = g_{\phi}(z)
\]

where \( \hat{x} \) is the reconstructed image, and the final layer uses a Conv2DTranspose layer with a sigmoid activation to output the reconstructed images.

The autoencoder was trained using the Adam optimizer \cite{kingma2014adam} with a learning rate of 0.001. The loss function used was Mean Squared Error (MSE), which measures the difference between the original image \( x \) and the reconstructed image \( \hat{x} \):

\begin{equation}
\text{MSE} = \frac{1}{n} \sum_{i=1}^{n} (x_i - \hat{x}_i)^2
\label{eq:mse}
\end{equation}

where \( n \) is the number of pixels in the image. The model was trained for 80 epochs with a batch size of 256.

\begin{figure}[htbp]
\centerline{\includegraphics[width=0.5\textwidth,height=\textheight,keepaspectratio]{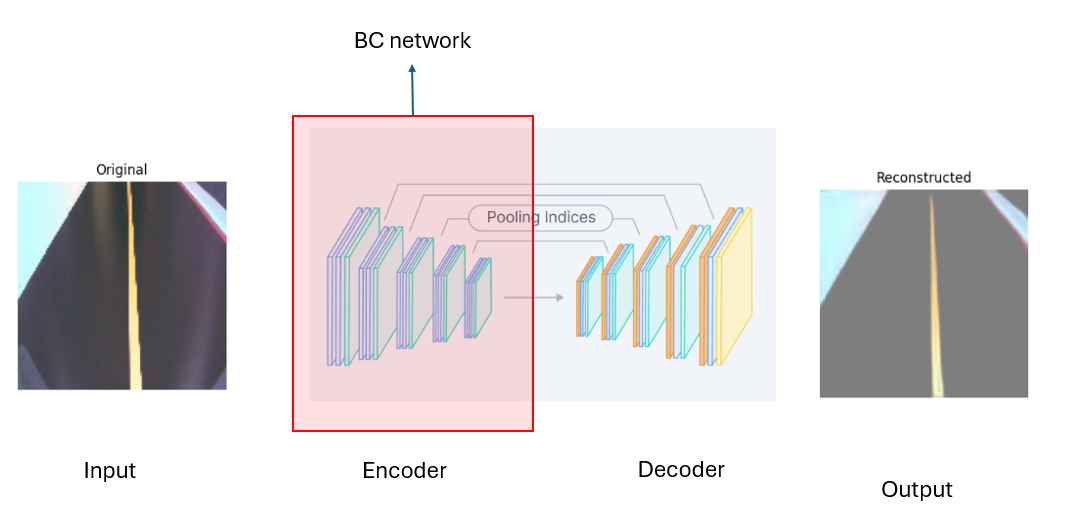}}
\caption{ Autoencoder network with encoded-decoded image result 8000 images with 80 epoch.After constructed image successfully, encoder network as in red highlighted part is used for BC}
\label{autoencoder_image}
\end{figure}

The performance of the autoencoder was visually evaluated by comparing the original and reconstructed images, as shown in Figure \ref{autoencoder_image}. The reconstructed images are quite similar to the input images, indicating that the network has successfully learned meaningful representations. After obtaining a good quality of reconstructed images, the trained encoder network was extracted for use in the behavioral cloning model.

\subsubsection{AutoBC}

\begin{figure}[hbp]
\centerline{\includegraphics[width=0.5\textwidth,height=\textheight,keepaspectratio]{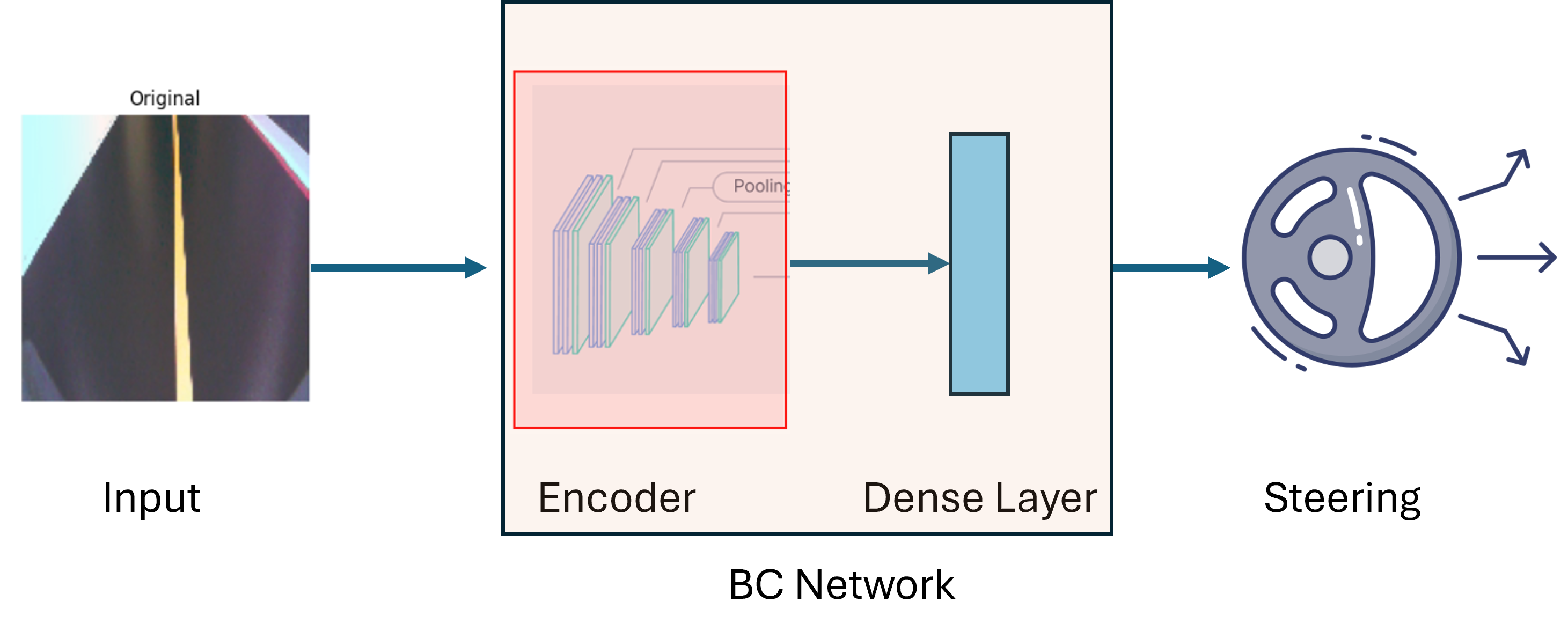}}
\caption{Autobc network:encoder network combined with dense layer to generate behaviour cloning network}
\label{autobc_network}
\end{figure}

The behavioral cloning model used the encoder network from the trained autoencoder \cite{chen2016variational} as shown in Figure \ref{autobc_network}. The encoder's output was flattened and passed through a dense layer with 128 ELU units and dropout for regularization. The final output layer was a dense layer with a linear activation to predict the steering angle.

The same preprocessing steps as the autoencoder \cite{chen2016variational} training were applied to ensure consistency. The BC model was trained using the Adam optimizer \cite{kingma2014adam} with a learning rate of 0.001, the same as the autoencoder. The MSE loss function was used. The model was trained for 50 epochs with a batch size of 64. Since we had already trained the encoder for 80 epochs, we trained it for an additional 50 epochs for the BC network.





\subsection{AutoBC with Spatial Attention}
\begin{figure}[htbp]
\centerline{\includegraphics[width=0.50\textwidth,height=\textheight,keepaspectratio]{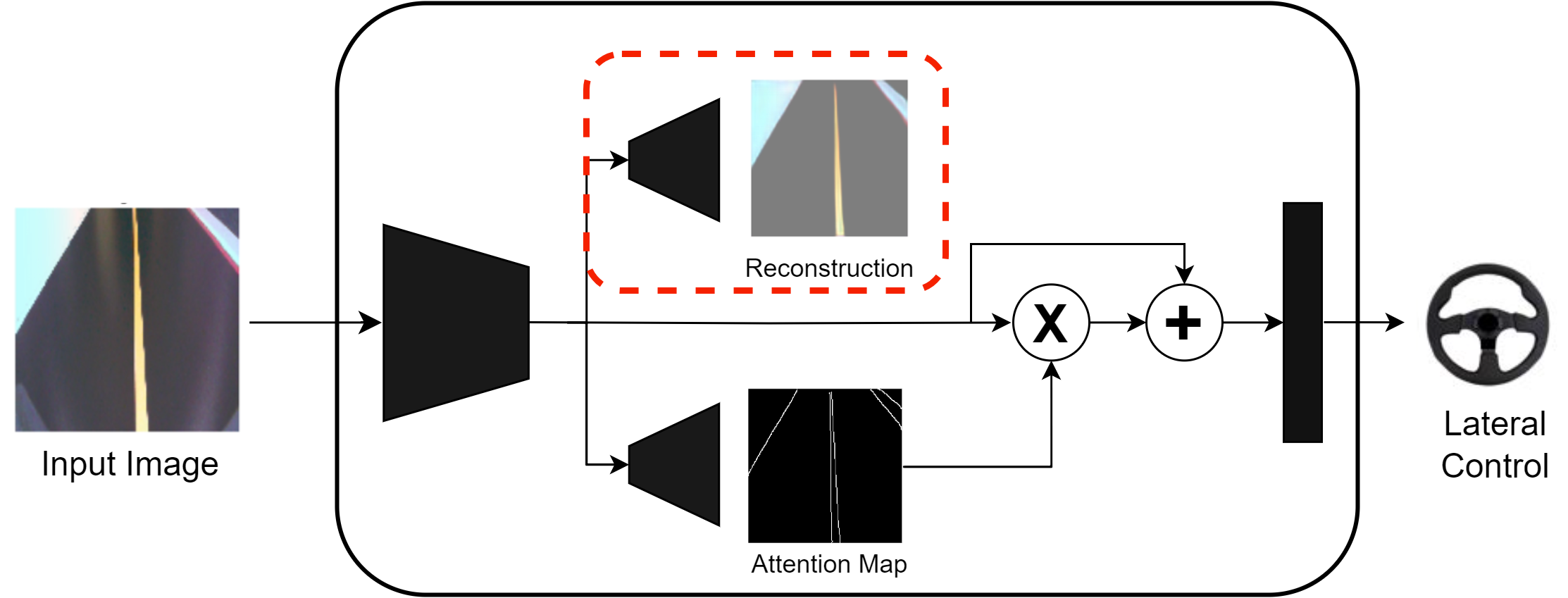}}
\caption{AutoBC with Spatial Attention architecture. The RGB image is passed through a dual-head autoencoder with a ResNet18 backbone, with the first head outputting the reconstruction of the original image and the second head outputting an attention mask that highlights the lane boundaries. The attention mask is then used to provide spatial attention to the original image features, to highlight the salient regions. These resulting features are sent to a dense layer to predict the steering angle.}
\label{Auto_BC_Spatial}
\end{figure}

The second proposed approach involves applying spatial attention to the image features of the AutoBC model. Spatial attention \cite{guo2022attention} is a mechanism that highlights salient regions of an image similar to how humans focus on important objects while performing tasks. As illustrated in Figure \ref{Auto_BC_Spatial}, the image observation (dimensions $3\times224\times224$) is first passed through the AutoBC encoder, producing a feature embedding that captures the essential characteristics of the original image. This embedding is then processed by two separate decoder streams: the AutoBC decoder, which reconstructs the original image but is removed during inference, and a second stream that generates a visual attention map. This attention map, with dimensions $1\times224\times224$, assigns values between $0$ and $1$, where areas containing lane markings have values close to $1$, and regions without lane markings have values closer to $0$. By highlighting lane boundaries, the visual attention map enables the model to focus on critical parts of the image, potentially improving the accuracy of steering predictions. This enhancement is crucial for the real-time application of AD systems, where precise lane detection is essential for safe navigation.

Spatial attention is applied to the feature map $\mathbf{Z}_{c} \in \mathbb{R}^{256 \times 56 \times56}$ from the penultimate layer of the AutoBC encoder. First, we perform an element-wise product between the visual attention map and the latent feature map $\mathbf{Z}_{c}$, producing an intermediate feature map $\mathbf{Z}_{att} \in \mathbb{R}^{256 \times 56 \times56}$. Next, an element-wise addition is performed between $\mathbf{Z}_{att}$ and the original feature map $\mathbf{Z}_{c}$, resulting in the final set of features with added spatial attention $\mathbf{Z}_{spatial} \in \mathbb{R}^{256 \times 56 \times56}$. These enhanced features are then flattened and passed through the final linear layer to output the lateral control. This method aims to improve the model's focus on relevant image regions, potentially enhancing the accuracy of steering predictions.

\subsection{Self-Supervised Transformer}

The third approach proposed for the BC is self-supervised pretrained (SP) vision transformer (ViT). The CNNs are good at understanding the local patterns within the image but ViTs perform better at global context understanding. The transformer models capture the long range dependencies and the relationship between image patches due to their multi-head self attention. As the ViTs are data hungry models by nature, we have employed a self-supervised pretarining technique to increase the in context learning of the transformer model.

\subsubsection{Pre-training} We are inspired from the Self-Supervised Vision Transformer approach proposed by \cite{atito2021sit}. The goal in our SP is to reconstruct a randomly masked image to enable the transformer model develop a pseudo label about the context. We utilise the grouped masked model learning approach and apply random masking to connected patches of the image. The masked image tokens are then passed through a multi-layer perceptron (MLP) \cite{kruse2022multi} to process the input features and transpose convolution is applied to recover the corrupted image patches considering the visible semantic concepts in the image.
The reconstruction loss used for our pretraining is defined as follows:

\begin{equation}
L_{\text{reconstruction}} = \frac{1}{N} \sum_{i=1}^{N} \| x_i - \hat{x}_i\|
\end{equation}

Where \( \lVert \cdot \rVert \) is the \( \ell_1 \) norm, \( x_i \) is the input image, \( \bar{x}_i \) is the reconstructed image and \( N \) is the batch size.

\subsubsection{Self-Supervised ViT for Steering Prediction}

We use the SP distilled vision transformer to fine-tune the model to predict steering angles for each image with the dataset that we collected. Similar to the previous approach, we assumed a constant throttle. The input images are split into 16x16 patches which are then tokenized, flattened and projected to pass through the transformer encoder. As we are predicting a single output, we have added a  MLP layer to map the 1000 dimensional image embedding output from the transformer into a single steering value. Our base transformer which is pretrained for image reconstruction learns to develop a global descriptor for the given image, the final MLP layer aims to learn the mapping from the the image embedding to the steering value.  

\begin{figure}[htbp]
\centerline{\includegraphics[width=0.45\textwidth,height=\textheight,keepaspectratio]{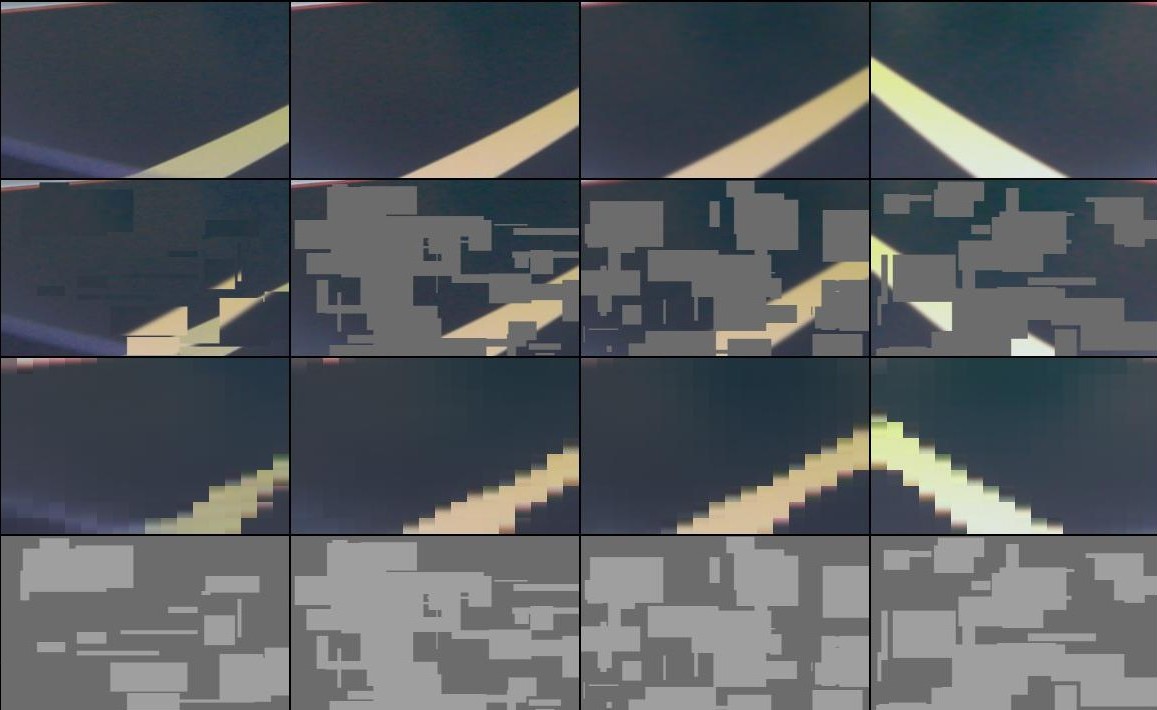}}
\caption{Samples from image reconstruction. Top row is the original image, second row is the masked image, third row is the reconstructed image and the last row shows the masking applied.}
\label{SP}
\end{figure}

\begin{figure}[htbp]
\centerline{\includegraphics[width=0.5\textwidth,height=\textheight,keepaspectratio]{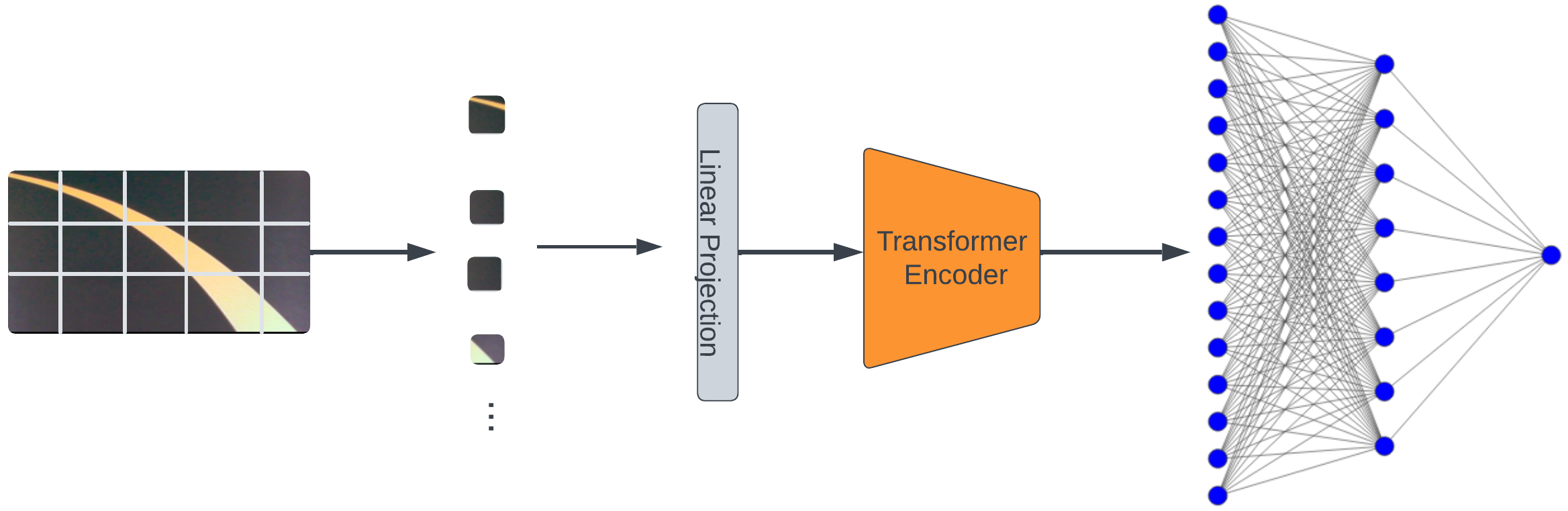}}
\caption{ViT pipeline}
\label{spvIT}
\end{figure}

\subsection{Vehicle Model}

\begin{figure}[htbp]
\centerline{\includegraphics[width=0.4\textwidth,height=\textheight,keepaspectratio]{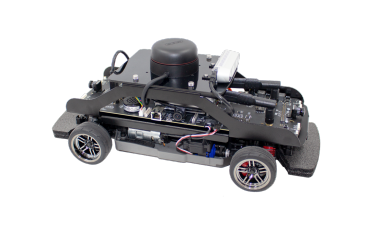}}
\caption{Scaled research vehicle:QCar }
\label{QCar_mage}
\end{figure}

The Qcar as shown in Figure \ref{QCar_mage}, which measures 20 cm wide, 19 cm tall, and 39 cm long, is equipped with sensors such as lidar, a camera, an IMU, and the Nvidia Jetson TX2 processor \cite{Qcar_product}. It has 4 cameras on each side and one RGB camera on top. To collect data we used front camera which is capable to capture road boundaries and beyond road with a wide angle specialities.

The TX2 acquires images from front camera process and transmit signals to PWM low-level signals to the vehicle's steering servo.

\subsubsection{Race Track}

\begin{figure*}[htbp]
\centerline{\includegraphics[width=\textwidth,height=\textheight,keepaspectratio]{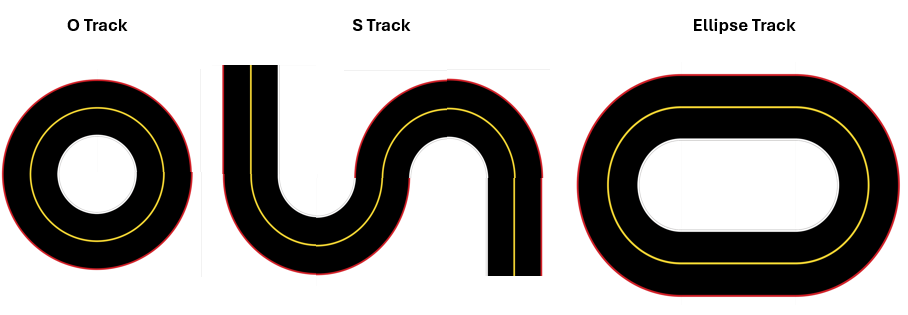}}
\caption{The racetrack map has been modified to test the methods under various road conditions. This version features dynamic road shapes and changing colors of the right and left road boundaries, enhancing the evaluation of the model's adaptability and performance.}
\label{Three_tracks}
\end{figure*}

We designed our in-house racetrack environment (as shown in Figure \ref{Three_tracks}) with different colors for each line: yellow for the center line to separate lanes, red for the outer border, and white for the inner border. This design allows the track to be used for multiple purposes, not only for imitation learning but also for rule-based methods. The lane widths are 25 cm to accommodate the scaled research vehicle width of 20 cm. The overall dimensions of the race map are 2.8 m by 1.8 m. It is designed as six pieces, each measuring 1 m by 1 m, making up the entire racetrack. This modular design also enables us to use it with different road configurations as shown in Figure \ref{Three_tracks}.

To train the model, we first optimized model structure and parameters. We initially started with a prebuilt CNN architecture, as shown in Figure \ref{CNN_nvidia}, and then modified it accordingly.

\subsection{Evaluation Metrics}

To evaluate the performance of the proposed models, we use three common metrics: Mean Absolute Error (MAE), Mean Squared Error (MSE), and Root Mean Squared Error (RMSE). Lower values across these metrics indicate better prediction accuracy.

The Mean Absolute Error (MAE) is defined as:

\begin{equation}
\text{MAE} = \frac{1}{n} \sum_{i=1}^{n} |y_i - \hat{y}_i|,
\label{eq:mae}
\end{equation}

where \(y_i\) represents the actual values, \(\hat{y}_i\) represents the predicted values, and \(n\) is the number of samples.

The Mean Squared Error (MSE) is defined as:

\begin{equation}
\text{MSE} = \frac{1}{n} \sum_{i=1}^{n} (y_i - \hat{y}_i)^2,
\label{eq:mse}
\end{equation}

where \(y_i\), \(\hat{y}_i\), and \(n\) are as defined above.

The Root Mean Squared Error (RMSE) is the square root of the MSE and is given by:

\begin{equation}
\text{RMSE} = \sqrt{\frac{1}{n} \sum_{i=1}^{n} (y_i - \hat{y}_i)^2},
\label{eq:rmse}
\end{equation}

where \(y_i\), \(\hat{y}_i\), and \(n\) are as defined above.

\section{Training}

\subsection{Dataset}

We drove the QCar on a designed racetrack for 2-3 hours and collected images, processing them by removing zero-velocity data. In the end, we obtained 20,000 (image, steering angle) tuples for training, corresponding to a total driving time of 6 minutes and 40 seconds. The steering angle, collected in radians, ranged from -0.5 to +0.5, with a maximum value of 0.5 radians (approximately 28.65 degrees to the right) and a minimum value of -0.5 radians (28.65 degrees to the left). The angle increments are highly detailed, varying up to three decimal places, as observed in the dataset Steering values shown in Table \ref{tab:dataset}. The images are collected in a folder, while the throttle and steering angle values are stored in a CSV file as shown at Table \ref{tab:dataset}.


\begin{table}[h]
\centering
\begin{tabular}{ccc}
\textbf{Frame}      & \textbf{Throttle}          & \textbf{Steering}       \\
\hline
Frame\_01.png & 0.065882349 & -0.001960814 \\
Frame\_02.png & 0.065882349 & -0.001960814 \\
Frame\_03.png & 0.049411762 & -0.001960814 \\
\end{tabular}
\caption{Dataset is collected as image frame number, aligned with throttle and steering angle}
\label{tab:dataset}
\end{table}

\subsubsection{Data Distribution}

\begin{figure}[htbp]
\centerline{\includegraphics[width=0.55\textwidth,height=\textheight,keepaspectratio]{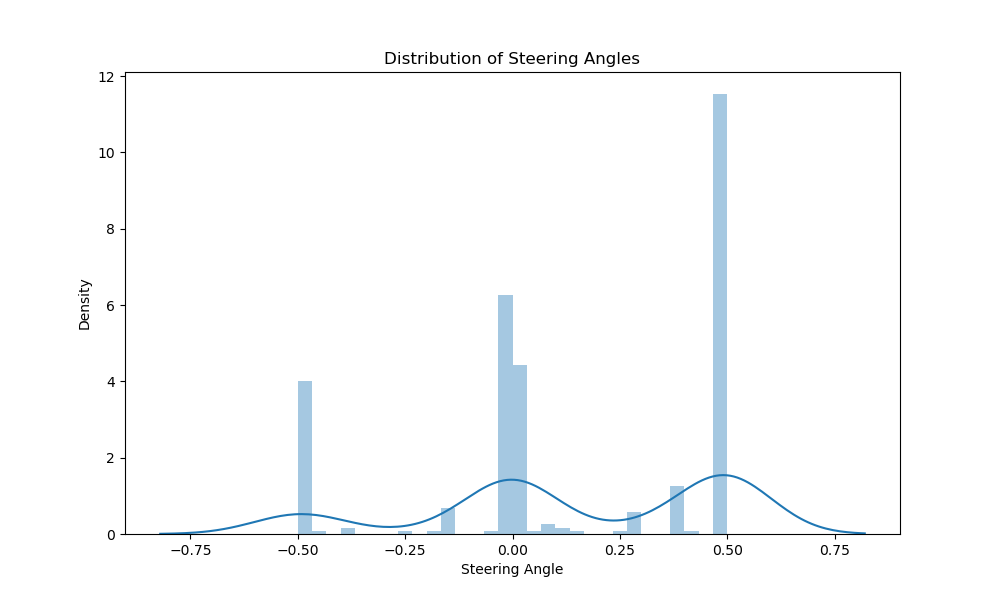}}
\caption{Steering angle distribution on  one sample of dataset}
\label{distribution}
\end{figure}

To understand the training process, we first investigated the data distribution within the model to examine the steering value distribution across the dataset. At the Figure \ref{distribution}, data distribution shows a significant imbalance, with high densities at specific steering angles (0.5, 0, and -0.5) and sparse coverage in between these values. This imbalance can lead to a biased behaviour cloning model that over-predicts the frequent steering angles and struggles with less-represented ones, potentially resulting in poor performance biased to learn the most frequent steering angle 0.5.

\subsection{Validation}

Based on the training and validation loss graphs, the validation loss increases and shows more fluctuations after approximately 25 epochs, indicating the start of overfitting. Although the training loss decreases, this is not correlated with improvements in the validation set, which is the primary indicator of model generalization. Therefore, training must stop at around 25 epochs, when the validation loss is at its lowest level before it begins to rise. Stopping at this point prevents overfitting, negatively impacting its performance on unseen data. The lowest validation loss achieves the optimal model performance, representing the best trade-off between bias and variance.

\begin{figure}[h]
\centerline{\includegraphics[width=0.5\textwidth,height=\textheight,keepaspectratio]{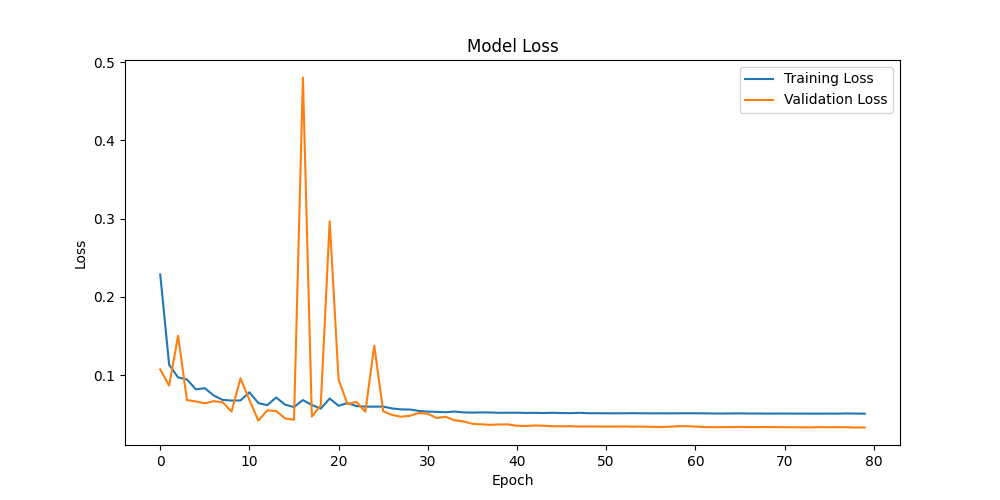}}
\caption{Training versus validation loss for autoencoder for epoch number 100}
\label{loss_80_epoch}
\end{figure}

\begin{figure}[h]
\centerline{\includegraphics[width=0.5\textwidth,height=\textheight,keepaspectratio]{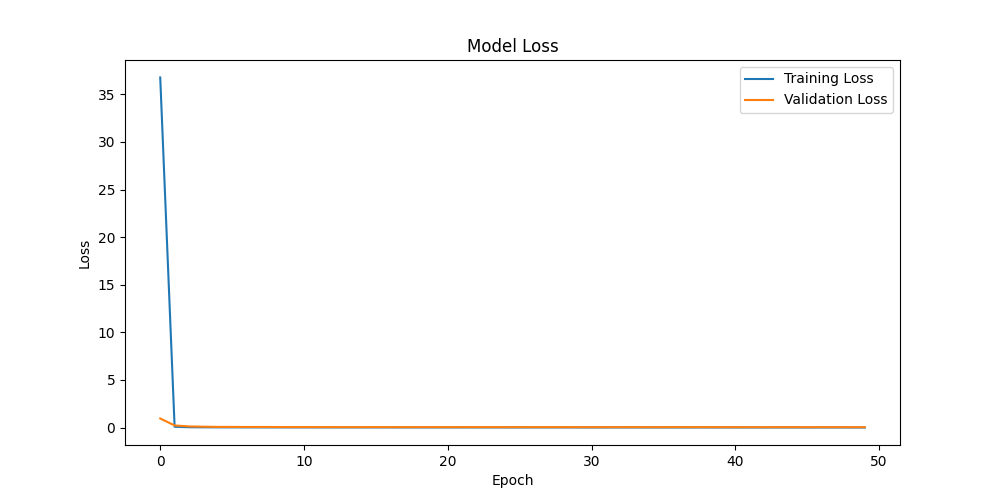}}
\caption{Training versus validation loss for BC for epoch 50}
\label{loss_50_epoch}
\end{figure}

\subsection{Data Augmentation}

Data augmentation was implemented to artificially expand the training dataset and improve model generalization by simulating various driving conditions. Several augmentation techniques were employed to enhance model generalization. Horizontal flipping was applied with a 50\% probability, which inverts the steering angle to help the model learn symmetrical driving behaviors for left and right turns. Vertical shifting alters the image's vertical position by up to 20\% to mimic changes in camera perspective due to vehicle pitch variations. However, this case was not part of the track environment, so it did not contribute to performance improvement. Darkening reduces pixel intensity by 50\% in random regions, simulating different lighting conditions such as shadows and varying light levels in an indoor setting, aiming to improve robustness in diverse lighting situations.

\begin{figure*}[h]
\centerline{\includegraphics[width=1\textwidth]{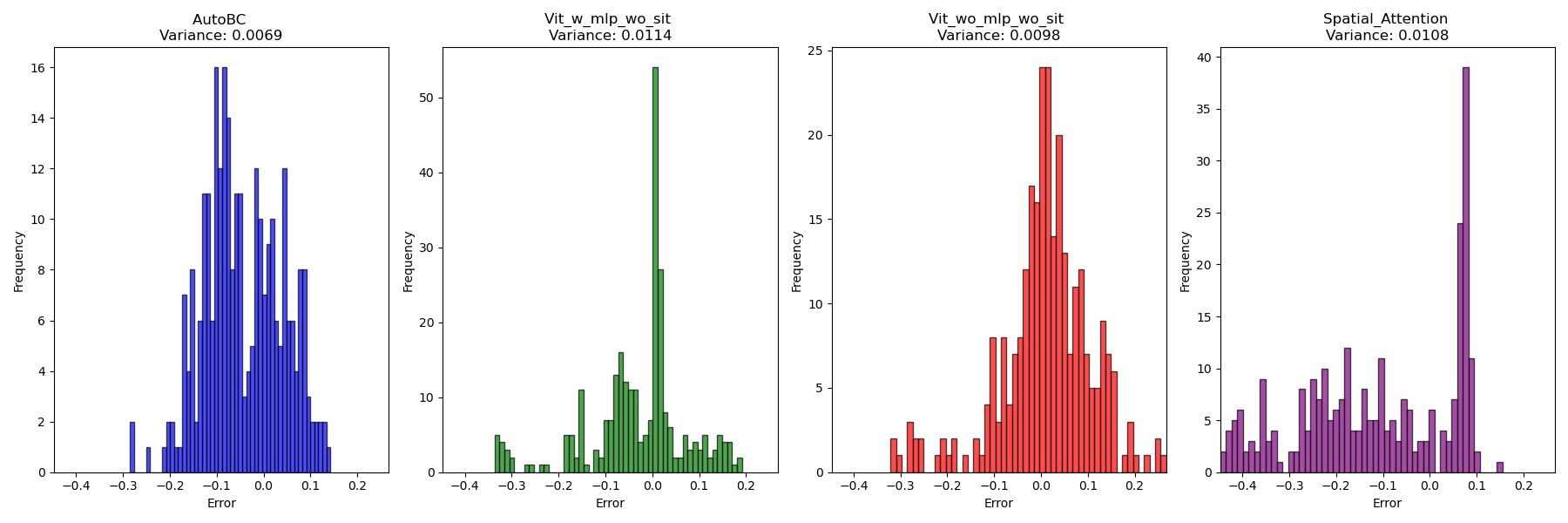}}
\caption{Error distribution and variance comparison of methods for steering angle prediction on the ellipse map}
\label{error_distribution}
\end{figure*}

The augmented images were normalized by scaling the pixel values between -0.5 and 0.5 to stabilize training. The augmentation process was integrated into the data pipeline during training using a generator function, ensuring that each batch of data was augmented on the fly, providing a more diverse dataset to the model.

\begin{figure*}[h]
\centerline{\includegraphics[width=\textwidth]{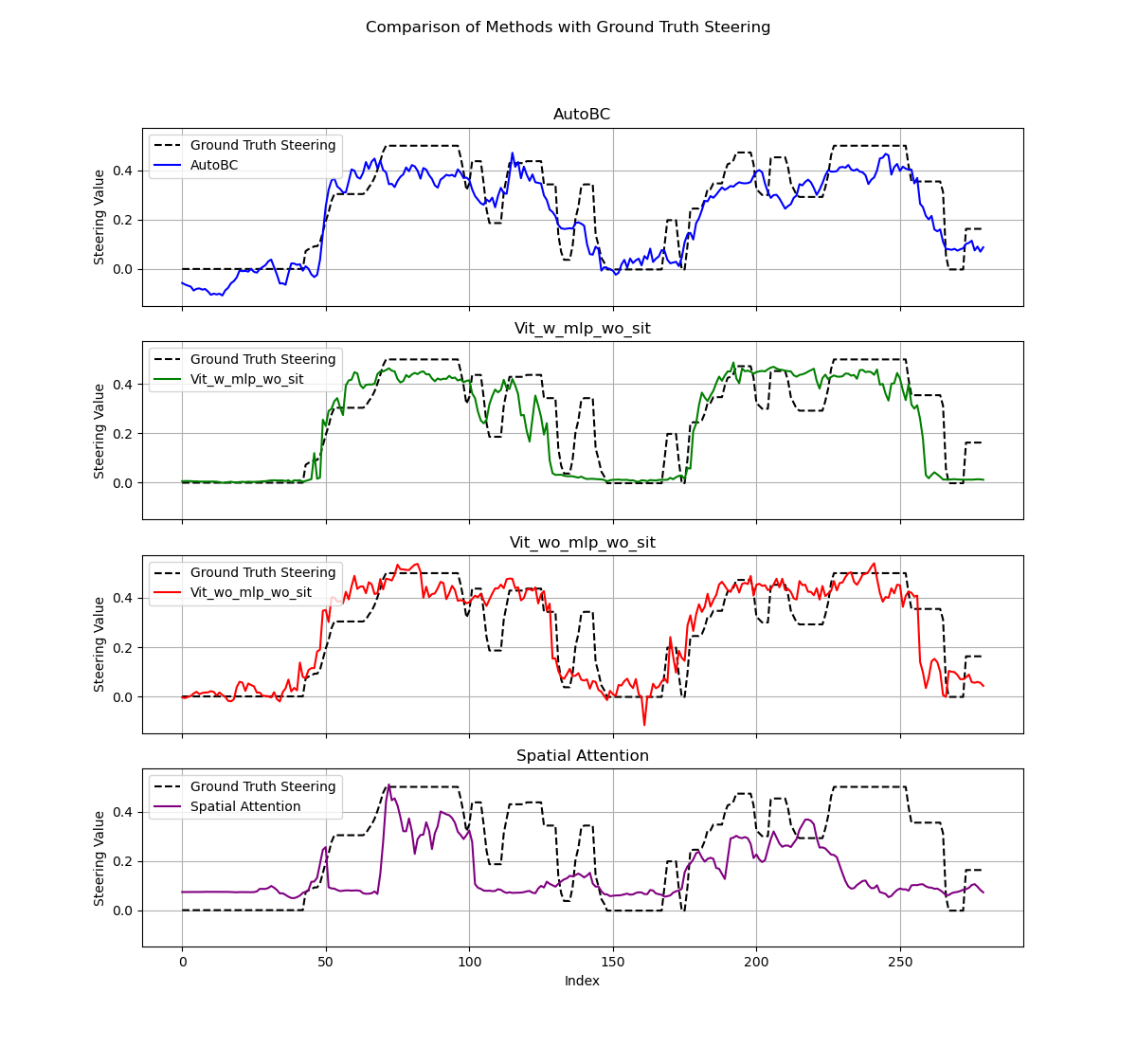}}
\caption{Comparison of methods prediction with ground truth steering angle for the ellipse map}
\label{ground_truth_vs_predicted}
\end{figure*}

To evaluate the impact of data augmentation on model performance, we trained the model with and without augmentation. The training loss with augmentation reached a smoothed value of 0.12 after 10 epochs, indicating stable learning. However, the model without augmentation showed much lower loss values, suggesting overfitting when augmentation was removed. While data augmentation is typically aimed at improving generalizability and model performance, our evaluation showed it did not help in our case. As noted in \cite{codevilla2019exploring}, augmentation is not always beneficial for improving performance, and another study \cite{du2019self} found that only a low augmentation ratio was beneficial; increasing augmentation deteriorated performance, so they used it minimally. Based on our evaluations and findings in the literature, we decided to remove augmentation.

\begin{figure}[h]
\centerline{\includegraphics[width=0.5\textwidth]{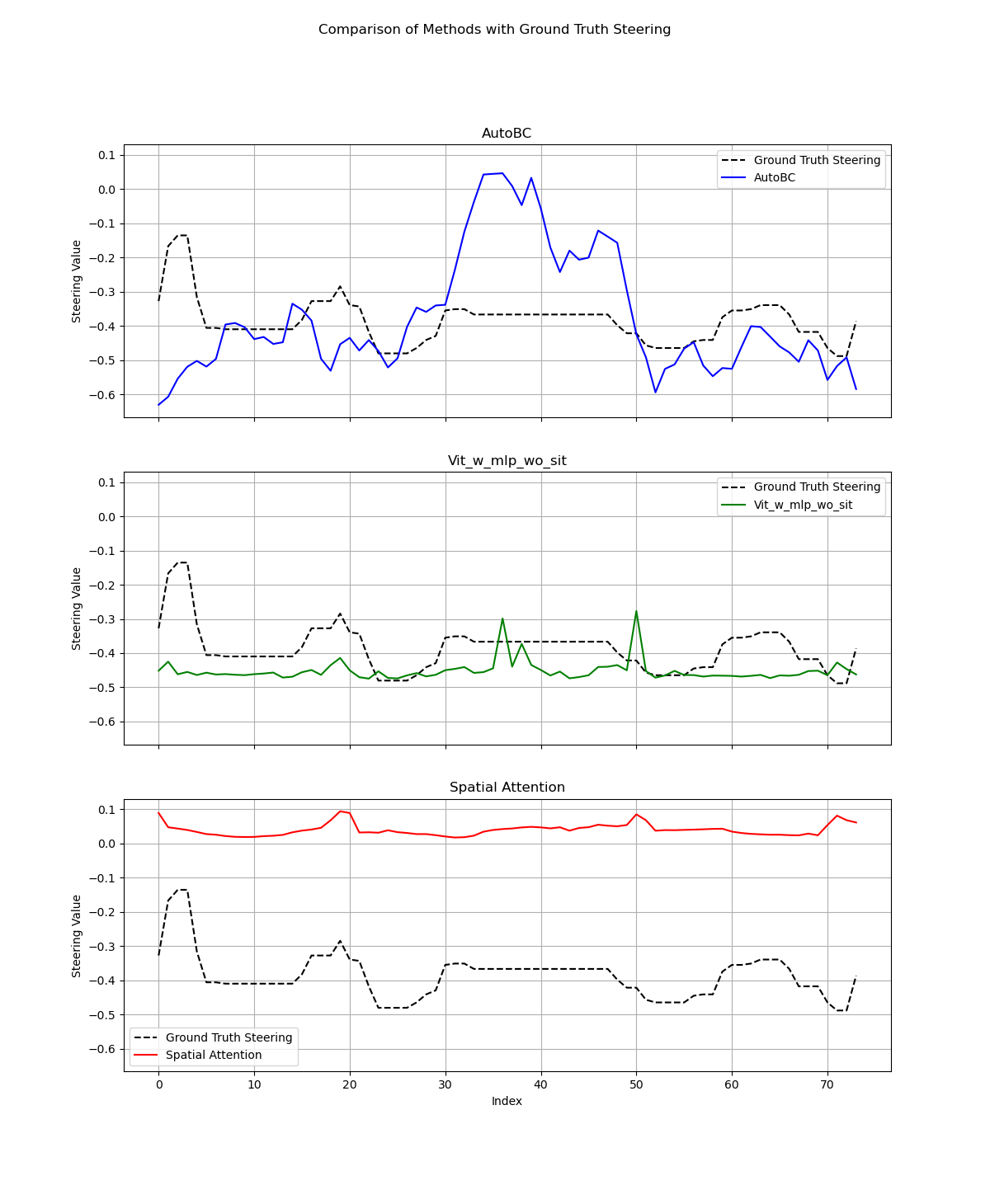}}
\caption{Comparison of methods prediction with ground truth steering angle for round 'O' map}
\label{ground_truth_vs_predicted_O_map}
\end{figure}

Subsequently, we collected 20 times more data to train the model. The number of epochs was increased to 80 for the autoencoder and 50 for the BC model, both without using augmentation.

\begin{figure*}[H]
\centerline{\includegraphics[width=1\textwidth]{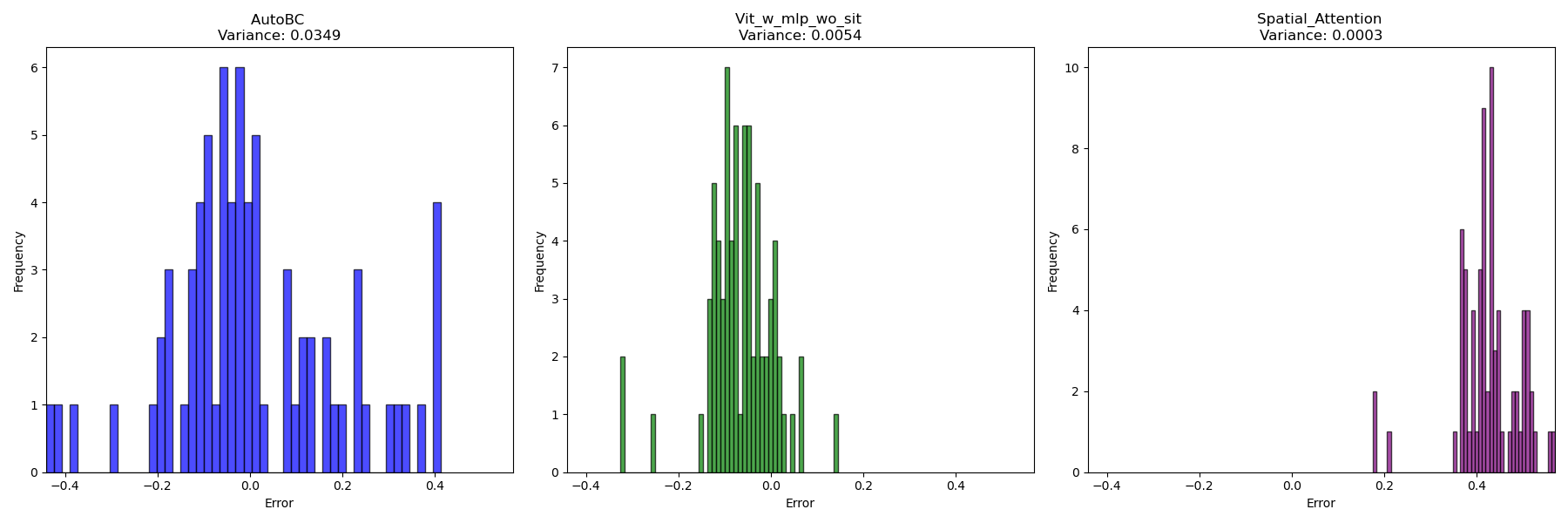}}
\caption{Error distribution and variance comparison of methods for steering angle prediction for O map}
\label{error_distribution_O_map}
\end{figure*}

\section{Experiments}

We collected datasets from three different track configurations: an elliptic map, a round 'O' map, and an 'S' map as shown in Figure \ref{Three_tracks}. For training the model, we used data exclusively from the elliptic map. Initially, we tested the trained model on a separate dataset also derived from the elliptic map to verify its performance.

To further evaluate the model's generalization capabilities, we tested it on unseen maps that differed from the training dataset. The round map (or 'O' map) was particularly valuable for testing the model's reliability, as it included only full right and close to full right steering angles, without any other steering data.

The 'S' map, which was unseen during training, provided a diverse testing environment with varying road shapes and changing road boundary colors. This setup was designed to assess the model's adaptability and its ability to focus on the main features of the road rather than relying on the colors of the road boundaries for navigation.


\section{Results}

The predicted and ground truth steering angles were saved to a CSV file for detailed analysis. To evaluate the model performance we checked the Mean Absolute Error (MAE), MSE, and the percentage of predictions within specific margins were calculated.

The prediction accuracy of the three methods is summarized in Table \ref{tab_prediction_accuracy}. We evaluate each method using three metrics: Mean Absolute Error (MAE), Mean Squared Error (MSE), and Root Mean Squared Error (RMSE). Lower values across these metrics indicate better performance in terms of prediction accuracy.

Mean Absolute Error (MAE) represents the average absolute difference between the predicted and actual steering values, with lower values indicating better accuracy. Among the three methods, ViT without MLP and SIT achieved the lowest MAE of 0.0795, indicating superior accuracy compared to AutoBC (0.0887) and ViT with MLP and without SIT (0.0828).

Mean Squared Error (MSE) represents the average squared difference between the predicted and actual steering values. ViT without MLP and without SIT also had the lowest MSE at 0.0117, reinforcing its strong performance. The MSE for AutoBC was slightly higher at 0.0119, while ViT with MLP and without SIT had the highest MSE of 0.0136.

Root Mean Squared Error (RMSE), which is the square root of the MSE and represents the standard deviation of the prediction errors. Consistent with the other metrics, ViT without MLP and without SIT showed the best performance with an RMSE of 0.1082. In contrast, ViT with MLP and without SIT recorded the highest RMSE of 0.1164, indicating it is the least accurate among the three methods.

\begin{table}[htb]
    \centering
    \begin{tabular}{cccc}
        \toprule
        Method & MAE & MSE & RMSE \\
        \midrule
        AutoBC & 0.0887 & 0.0119 & 0.1091 \\
        ViT with MLP without SIT & 0.0828 & 0.0136 & 0.1164 \\
        ViT without MLP without SIT & 0.0795 & 0.0117 & 0.1082 \\
        \bottomrule
    \end{tabular}
    \caption{Prediction Accuracy Metrics for Different Methods}
    \label{tab_prediction_accuracy}
\end{table}

Variance of methods tested for ellipse map, as shown in Figure \ref{error_distribution} shows the reliability of methods. AutoBC, implies more consistent and reliable predictions as it has lower variance as a value 0.0069 comparing other methods. ViT with MLP and without SIT has higher variance as value 0.0114 in methods, indicates more variability in the prediction errors, suggesting less stable predictions. ViT without MLP and without SIT has value of 0.0098, suggests a balance between the other two methods, with moderate consistency and reliability. Lastly spatial attention model has a value of 0.0108. Variance helps in assessing not just the accuracy but also the reliability of the prediction methods. AutoBC showed lower variance notifying model's predictions are consistently close to the actual values.

\begin{table}[htb]
    \centering
    \begin{tabular}{lc}
        \toprule
        \textbf{Margin} & \textbf{Percentage} \\ 
        \midrule
        Within 0.1 radians (5.73 degrees) & 61.25\% \\ 
        Within 0.2 radians (11.46 degrees) & 95.00\% \\ 
        Within 0.3 radians (17.19 degrees) & 99.64\% \\ 
        \bottomrule
    \end{tabular}
    \caption{Percentage of predictions within specific margins}
    \label{tab:prediction_margins}
\end{table}

As shown in Table \ref{tab:prediction_margins}, 61.25\% of the predictions fall within a 0.1 radian (5.73 degrees) error margin. For a 0.2 radian (11.46 degrees) margin, 95.00\% of the predictions are within this range. Almost all predictions, 99.64\%, are within a 0.3 radian (17.19 degrees) margin, which was the maximum range of error observed across all test predictions. These percentages indicate that a substantial proportion of the model's predictions are very close to the actual steering angles, demonstrating the model's reliability.

ViT without MLP and without SIT consistently shows the best performance across all three metrics (MAE, MSE, RMSE), indicating it is the most accurate method for predicting steering values. ViT with MLP and without SIT has the highest errors, suggesting it is the least accurate among the three methods. AutoBC performs moderately well, better than ViT with MLP and without SIT but not as well as ViT without MLP and without SIT.

We tested three methods on a unseed round 'O' map, considering all methods were trained only on an ellipse map. The round map showed that all methods deteriorated in performance and accuracy, exhibiting worse performance than on the ellipse map, as shown in Figure \ref{ground_truth_vs_predicted_O_map}. The ViT model was tested only with MLP and without SIT. When compared to AutoBC and the AutoBC spatial attention model, it showed better performance throughout the test track. AutoBC showed close prediction at some points, but the AutoBC spatial attention version completely lost prediction accuracy, showing a huge margin from the ground truth.

The AutoBC model with spatial attention performed poorly at predicting steering, likely due to the objects highlighted in the attention mask. Highlighting only the lane boundaries may have been insufficient because they occupy minimal space in the image. A more effective approach might be to apply spatial attention to the drivable road, which occupies a larger portion of the image. Conversely, spatial attention might be better suited for highlighting specific objects, such as pedestrians, when addressing collision avoidance. This suggests that the effectiveness of spatial attention depends on the specific task and the prominent features required for accurate prediction.

\section{Conclusion}\label{conc}

We investigated end-to-end autonomous driving for a racetrack and compared state-of-the-art methods. We implemented autoencoder-based behavioral cloning, vision transformer, and spatial attention-based networks. The dataset was collected using a scaled research vehicle and tested on different track setups. The evaluation of the autoencoder-based behavioral cloning model revealed promising results on the ellipse track but showed deteriorated performance on the unseen map. The AutoBC spatial attention model showed worse performance on both tracks. On the other hand, the ViT model demonstrated better performance and generalizability on both the trained and unseen maps. These results indicate that the ViT model is capable of making precise steering angle predictions, with a high percentage of predictions falling within acceptable margins. 

Future work could focus on reducing error margins and increasing model robustness through architectural enhancements and hyperparameter tuning. Additionally, evaluating the AutoBC method on diverse datasets, such as RoboBus \cite{varisteas2021robobus}, would help assess its generalizability across different dataset. Another promising direction is incorporating retrieval augmentation of images, which maps images directly to steering actions. This approach could be compared with current methods to evaluate performance.

\bibliographystyle{ieeetr}
\bibliography{sample}
\vspace{12pt}

\end{document}